\documentclass[12pt]{article}
\usepackage{arxiv}

\usepackage{titlecaps}

\usepackage[numbers,sort&compress]{natbib}

\usepackage{xcolor}     %
\usepackage{graphicx}   %

\usepackage[inline]{enumitem} %
\usepackage{booktabs}       %
\usepackage{multirow, array, threeparttable}
\usepackage[font=normalsize,skip=4pt]{caption}
\usepackage{subcaption}
\usepackage{algorithm,algpseudocode}

\algnewcommand\algorithmicinput{\textbf{Input:}}
\algnewcommand\algorithmicoutput{\textbf{Output:}}
\algnewcommand\algorithmicnote{\textbf{Note:}}
\algnewcommand\Input{\item[\algorithmicinput]}%
\algnewcommand\Output{\item[\algorithmicoutput]}%
\algnewcommand\Note{\item[\algorithmicnote]}%

\usepackage[utf8]{inputenc} %
\usepackage[T1]{fontenc}    %
\usepackage{microtype}      %
\usepackage{url}            %
\usepackage{doi}
\usepackage{nicefrac}       %
\usepackage{amsfonts}       %

\usepackage{amsmath}
\usepackage{amssymb}
\usepackage{mathtools}
\usepackage{amsthm}
\usepackage{amsopn} %
\DeclareMathOperator{\diag}{diag}
\newcommand{\diff}{\mathrm{d}}  %
\usepackage{bm}

\usepackage{hyperref}
\hypersetup{
    colorlinks,
    linkcolor={blue!60!black},
    citecolor={red!60!black},
    urlcolor={blue!80!black}
}
\usepackage[capitalize,nameinlink]{cleveref} %

\usepackage{siunitx} %
\makeatletter
\renewcommand{\paragraph}{%
  \@startsection{paragraph}{4}{\z@}%
  {0.25\baselineskip}  %
  {-0.25em}            %
  {\normalfont\normalsize\bfseries}%
}
\makeatother

\usepackage{hyphenat} %
\title{Multi-fidelity Machine Learning for Uncertainty Quantification and Optimization}
\renewcommand{\shorttitle}{Multi-fidelity ML for UQ \& Optimization}

\author{
{\hspace{1mm}Ruda Zhang}\thanks{Corresponding author.} \\
	University of Houston\\
	\texttt{rudaz@uh.edu} \\
        \And
{\hspace{1mm}Negin Alemazkoor}\\
	University of Virginia\\
	\texttt{na7fp@virginia.edu} \\
}

\newcommand{\matern}{{Mat\'ern }} %

\date{}

\renewcommand{\headeright}{ }

\begin{document}
\maketitle

\begin{abstract}
In system analysis and design optimization,
multiple computational models are typically available to represent a given physical system.
These models can be broadly classified as high-fidelity models,
which provide highly accurate predictions but require significant computational resources,
and low-fidelity models, which are computationally efficient but less accurate.
Multi-fidelity methods integrate high- and low-fidelity models to balance computational cost and predictive accuracy.
This perspective paper provides an in-depth overview of the emerging field of machine learning-based
multi-fidelity methods, with a particular emphasis on uncertainty quantification and optimization.
For uncertainty quantification, a particular focus is on multi-fidelity graph neural networks,
compared with multi-fidelity polynomial chaos expansion.
For optimization, our emphasis is on multi-fidelity Bayesian optimization,
offering a unified perspective on multi-fidelity priors and proposing an application strategy
when the objective function is an integral or a weighted sum.
We highlight the current state of the art, identify critical gaps in the literature,
and outline key research opportunities in this evolving field.\footnote[1]{This is the accepted version of the following article:
Zhang, R. \& Alemazkoor, N.
Multi-fidelity Machine Learning for Uncertainty Quantification and Optimization.
\textit{Journal of Machine Learning for Modeling and Computing}, Vol. 5, No. 4, pp. 77--94, (2024),
\url{https://doi.org/10.1615/JMachLearnModelComput.2024055786}.
This version is made available under the terms of the Green Open Access Agreement.
}
\end{abstract}

\keywords{multi-fidelity modeling \and uncertainty quantification \and Bayesian optimization}

\section{Introduction}
\label{sec:intro}

When studying a physical system, analysts often have access to multiple computational models.
These models are typically classified as either high-fidelity or low-fidelity,
depending on their predictive accuracy.
High-fidelity models (HFMs) offer precise predictions of the system's behavior, meeting specific accuracy metrics,
but they are computationally demanding.
This computational demand is driven by the need for fine mesh resolutions
and small time steps to ensure both numerical stability and accuracy.
On the other hand, low-fidelity models (LFMs) trade off some accuracy for greater computational efficiency.
These models are often derived through techniques like mesh coarsening, simplified physics, or reduced-order modeling,
resulting in less computationally expensive simulations.

While relying exclusively on LFMs can reduce computational time,
it risks producing myopic results due to their inherent lower accuracy.
Conversely, replying solely on HFMs for analysis, especially in high dimensional spaces,
can be computationally prohibitive.
This limitation may hinder timely decision-making, which is critical for optimal system operation and management.
Multi-fidelity (MF) methods aim to strike a balance between LFMs and HFMs,
combining computational efficiency with predictive accuracy to enhance decision-making in complex systems.

The literature on MF methods is extensive, encompassing numerous strategies for integrating LFMs and HFMs
across diverse scientific and engineering applications.
\citet{peherstorfer2018survey} offers a comprehensive survey of MF methods
in uncertainty quantification (UQ), inference, and optimization.
Since then, there has been a significant shift in the literature
towards integrating machine learning (ML) with MF methods.

In this work, we provide an overview of multi-fidelity machine learning methods,
focusing particularly on surrogate modeling techniques
and applications in uncertainty quantification and optimization.
Specifically, we discuss the gaps in the literature, identify research opportunities,
and suggest future directions for this evolving field.

\Cref{sec:mfuq} discusses MF surrogates for uncertainty quantification,
with particular focuses on MF polynomial chaos expansion and graph neural networks.
\Cref{sec:mfo} discusses MF optimization and, in particular, MF Bayesian optimization.
We provide a unified perspective on MF priors and propose a practical application in quadrature-type objective functions.
In each section, we highlight challenges and propose future directions.
\Cref{sec:conclusion} concludes this article.

\section{Multi-fidelity Surrogates and Uncertainty Quantification}
\label{sec:mfuq}

Accurately simulating complex physical systems requires significant computational resources and storage,
even with advanced parallel processing techniques.
The high-dimensional nature of these systems, combined with uncertainty in various parameter values,
makes it impractical to rely on deterministic assumptions for a single simulation run.
Uncertainty quantification (UQ) methods are essential for
evaluating how random input variables propagate through the system
and quantifying the resulting output uncertainties.
Specifically, let the random input be represented by ${\bm \Xi}$, with a probability density function $\rho$.
The high-fidelity model then maps these inputs to the uncertain system output,
represented as: $f_{\text{hi}}({\bm \Xi}) = Y$.
The goal of UQ is to estimate the statistics of the output $Y$, including its expectation and variance.

Monte Carlo (MC) simulation---the most commonly used approach to UQ---draws
$m$ independent and identically distributed samples of the random inputs
and estimates the expectation of the output as the sample average:
\begin{equation}
    \overline{Y}_{m}^{\text{hi}} := \frac{1}{m}\sum_{i=1}^{m}f_{\text{hi}}({\bm \Xi}_i)
    \; \approx \; \mathbb{E}[Y] := \int f_{\text{hi}}({\bm \xi}) \rho({\bm \xi}) \diff {\bm \xi}.
\end{equation}
The mean square error of this estimator is:
\begin{equation}
  e_\text{MSE}\big(\overline{Y}_{m}^{\text{hi}}\big) = \frac{1}{m} \mathbb{V}[Y],
\end{equation}
where $\mathbb{V}[Y]$ is the variance of $Y$.
The main advantages of MC methods for UQ are their simplicity, ease of implementation,
and a convergence rate that is independent of function smoothness and dimensionality of the random input.
However, the main disadvantage is its slow convergence rate of $\mathcal{O}(m^{-{1}/{2}})$ in sample size.

Given that the variance of $f_{\text{hi}}({\bm \Xi})$ directly impacts the convergence rate,
several variants of MC aim at facilitating the estimation of $\overline{Y}_{m}^{\text{hi}}$
through integrating an alternative function with a smaller variance.
Such methods include importance sampling \citep{tokdar2010importance}, stratified sampling\citep{shields2015refined},
and conditional MC sampling \citep{fu2012conditional}.
Still, all these variants require a significant number of HFM evaluations.

More recently, MF MC approaches have been proposed that leverage the low computational cost of generating data from LFMs.
By strategically integrating LFMs and HFMs,
MF MC sampling reduces the number of costly high-fidelity evaluations required
while still capturing the system's essential features.
This reduction is typically achieved through a control variate approach,
where discrepancies between LFMs and HFMs are exploited to improve the accuracy of the estimates.
\citet{peherstorfer2018survey} provide a comprehensive overview of MF MC approaches and their variants,
so we do not review these methods here.
Instead, we focus on MF approaches within an alternative class of UQ methods: surrogate modeling.

Surrogate models can accurately estimate the statistics of the system's response with far fewer evaluations than MC sampling,
particularly when the dimensionality of the random input is small to moderate.
Additionally, surrogate models can capture the underlying input--output relationships more effectively,
facilitating tasks such as prediction, sensitivity analysis, dimension reduction, and reliability analysis.
Surrogates can generally be classified into two main categories:
(1) analytical surrogates and (2) machine learning-based surrogates.
Analytical surrogates include methods like polynomial chaos expansion (PCE) and reduced-order models.
These approaches rely on well-defined mathematical frameworks to model the input--output relationship.
Machine learning-based surrogates include methods such as deep neural networks (DNNs),
graph neural networks (GNNs), support vector machines (SVMs), and Gaussian processes (GPs) \citep{Jakeman2022}.
These models leverage data-driven techniques to learn complex patterns and relationships from the data.
The distinction between these two categories is not always clear-cut;
hybrid surrogates can be constructed, for example,
by combining Gaussian processes with reduced-order models \citep{ZhangRD2022gps}.
In this section, for brevity,
we compare PCE and GNN as representative examples of analytical and machine learning-based surrogates,
and we discuss the state of the art, challenges, and future directions of MF ML techniques
in enhancing surrogate modeling.

\subsection{Multi-fidelity Polynomial Chaos Expansion}

Let the system input $\bm \Xi$ be a $d$-dimensional vector of independent random variables,
where each random variable $\Xi_i$ has a support $I_{ \Xi_i}$
and a marginal probability density function $\rho_i: I_{\Xi_i} \mapsto \mathbb{R}^+$.
Then $\bm \Xi$ is supported on the product domain $I_{\bm \Xi} = \prod_{i=1}^d I_{ \Xi_i}$,
with a joint probability density function $\rho(\bm \Xi) = \prod_{i=1}^{d}\rho_i(\Xi_i)$.
Any square-integrable random output, denoted by $u(\bm \Xi): I_{\bm \Xi} \mapsto \mathbb{R}$,
can be represented as:
\begin{equation}\label{eq:repUasSum}
u(\bm{\Xi}) = \sum_{\bm \alpha \in \mathbb{N}_0^d} c_{\bm \alpha } \psi_{\bm \alpha}(\bm\Xi), 
\end{equation}
where $\bm \alpha=(\alpha_1, ..., \alpha_d)$ is a multi-index,
$c_{\bm \alpha }$ are the PCE coefficients to be estimated,
and $\{ \psi_{\bm \alpha} \}_{\bm \alpha \in \mathbb{N}_0^d}$ is the set of basis functions
that satisfy the orthogonality condition:
\begin{equation} \label{eq:orogonality}
\int_{I_{\bm \Xi}} \psi_{\bm m}(\bm \xi) \psi_{\bm n}(\bm \xi) \rho({\bm \xi}) \diff{\bm \xi}
= \delta_{\bm {mn}}, \quad \bm{m}, \bm{n} \in \mathbb{N}_0^d.
\end{equation}
In practice, a $p$-th order truncation of the expansion in \cref{eq:repUasSum}
is used to approximate $u(\bm \Xi)$:
\begin{equation}\label{eq:Approximation}
u(\bm\Xi) \approx u_{p}(\bm\Xi) = \sum_{\bm \alpha \in \Lambda_{d, p}} c_{\bm \alpha } \psi_{\bm \alpha}(\bm \Xi),
 \end{equation}
where $\Lambda_{d, p}$ is the set of multi-indices with order equal to or less than $p$,
\begin{equation}\label{eq:Truncation}
\Lambda_{d, p} := \{\bm \alpha \in \mathbb{N}_0^d : \Vert \bm \alpha \Vert_1 \le p \}.
\end{equation}
The cardinality of $\Lambda_{d, p}$, denoted by $K$, can be calculated as:
\begin{equation}\label{eq:NumOfbasis}
K := \vert\Lambda_{d, p}\vert = \frac{(p+d)!}{p! \ d!}.
\end{equation}

Different non-intrusive approaches such as spectral projection \citep{le2010spectral,conrad2013adaptive},
sparse grid interpolation \citep{barthelmann2000high,buzzard2012global},
and least squares \citep{guo2020constructing,hampton2015coherence} have been used to construct the PCE approximation.
However, all these approaches suffer from the curse of dimensionality.
Specifically, the number of required sample points for accurate PCE surrogate construction increases drastically
with the dimensionality of the random input.
Depending on the non-intrusive method used for PCE construction,
various techniques have been introduced to alleviate the challenge of the cures of dimensionality.
Examples include adaptive methods for sparse grids \citep{hegland2002adaptive,brumm2017using},
basis selection \citep{thapa2020adaptive,blatman2011adaptive},
and optimal sampling \citep{hampton2015coherence,shin2016near, taghizadeh2024improving, shin2016nonadaptive}
for improving the accuracy of PCE construction using the least-square method.
Given that approximated PCEs for many high-dimensional problems are often highly sparse, compressive sampling,
a technique that allows for fewer samples than the number of unknown coefficients,
has been widely used for PCE construction \citep{hampton2015compressive}.
The accuracy of sparse PCE has been shown to depend heavily on the incoherence properties of the measurement matrix,
specifically the Vandermonde-like matrix formed by evaluating orthogonal polynomials at the sample points.
Consequently, several efforts have focused on improving the incoherence properties of the measurement matrix
to enhance approximation accuracy, utilizing techniques such as optimal sampling \citep{alemazkoor2018near},
adaptive basis selection \citep{jakeman2015enhancing}, dimension reduction \citep{alemazkoor2017divide},
and preconditioning \citep{alemazkoor2018preconditioning} in under-determined settings.

Among all methods proposed to facilitate PCE construction, multi-fidelity methods have proven to be the most efficient,
when informative low-fidelity simulations of the complex physical system of interest are available.
Multi-fidelity PCE was first introduced by \citet{ng2012multifidelity}.
Specifically, the authors applied the correction response surface approach by \citet{vitali2002multi}
to PCEs using stochastic collocation.
Their proposed method utilizes different levels of sparse grids,
where the lower-level sparse grid is a subset of the higher-level grid and functions as a corrector for it.
The low- and high-level sparse grids are used to construct the PCE approximation for the LFM
and the discrepancy between the LFM and the HFMs, respectively.
\citet{bryson2017all} extended this approach to simultaneously estimate the coefficients of
the low-fidelity PCE and the correction terms, further improving the accuracy of surrogate estimation.
The main disadvantage of the approach proposed by \citet{ng2012multifidelity} and \citet{bryson2017all} is that
the number of high-fidelity samples depends on the level of the sparse grid and cannot be arbitrary.
This can be especially problematic when a limited computational budget is available.
Additionally, the locations of the sample points are determined by the sparse grid,
making it difficult to integrate an existing set of random simulations with this approach. 

\citet{palar2018global} proposed the first regression-based multi-fidelity approaches for PCE construction,
where regression is used to construct PCE approximations for the LFM and the discrepancy between the HFM and LFM.
Regression allows an arbitrary number and location of samples, as opposed to sparse grids,
which is more favorable in cases where existing samples are available or there are computational budget restrictions.
Given that, especially in high-dimensional problems, several PCE coefficients are either zero or have insignificant values,
\citet{cheng2019multi} proposed an iterative approach that assigns more weight to those basis functions
identified as more significant using the low-fidelity samples.
Based on the same motivation, \citet{salehi2018efficient} performed $L_1$ minimization on low-fidelity samples
to identify the significant PCE coefficients, consequently minimizing the number of high-fidelity samples
needed to construct the PCE approximation for the discrepancy between the HFM and the LFM.
\citet{salehi2018efficient} set the number of high-fidelity samples to be twice the cardinality
of the identified significant coefficients.
This is based on the accepted rule of thumb that accurate coefficient estimation using least squares
requires an oversampling rate of 1.5 to 3 \citep{shin2016nonadaptive}.

Most recently, \citet{alemazkoor2022multi} proposed a multi-fidelity optimal sampling scheme
that successfully reduces the number of required high-fidelity simulations for accurate PCE construction
to be equal to the number of identified significant basis functions using the LFM.
Specifically, this is done by searching a large pool of low-fidelity samples
and systematically selecting the most high-yield ones from the pool and running the HFM at those sample locations. 

PCEs and MF PCEs have been used as surrogates for HFMs in several applications, including but not limited to 
structure property estimation \citep{alemazkoor2022multi},
aerodynamic analysis and shape optimization of aircraft \citep{zhao2021adaptive},
and power system analysis \citep{alemazkoor2020fast}.
PCE is popular due to its analytical formulation, simplicity of construction, and the ability to directly compute
the mean, variance, and Sobol indices of the QoI from the PCE coefficients \citep{sudret2008global}.
Recent advancements in MF modeling have significantly reduced the need
for a large number of high-fidelity samples in PCE construction.
However, there is still room for improvement, such as developing optimal sampling strategies
that incorporate more than two fidelity levels or effectively utilize existing data of unknown fidelity.
Despite these potential enhancements, the literature appears to be shifting towards machine learning-based surrogates.
This shift is driven by the increasing complexity of modern engineering and scientific problems,
where the assumptions and limitations of analytical surrogates like PCE may not hold.
For example, each PCE surrogate estimates a single quantity of interest (QoI),
while multi-output machine learning models can simultaneously be trained to estimate or predict several QoIs.
Consequently, machine learning-based surrogates offer greater flexibility and scalability,
enabling them to model highly non-linear and complex systems more effectively.
Therefore, we next review advances in MF modeling for an increasingly popular type of machine learning surrogate,
namely graph neural networks.

\subsection{Multi-fidelity Graph Neural Networks}

One of the main advantages of Graph Neural Networks (GNNs) over analytical surrogates
and other conventional machine learning approaches is that GNNs account for the system topology,
eliminating the need to retrain a surrogate every time there is a minor change in the system's topology.
We start with a brief overview of how GNNs incorporate the topology of the system,
and then survey the applications of multi-fidelity GNNs.

\subsubsection{Basics of GNNs}

Consider a graph $G=(V,E)$, where $V$ denotes the set of nodes and $E$ denotes the set of edges.
Each link is denoted by $(u,v) \in E$, with $u$ and $v$ being its two end nodes.
Let $|V|$ denote the number of nodes and $|F_v|$ denote the dimension of features for each node.
Similarly, define $|E|$ and $|F_e|$ for edges.
GNNs map node features $X_n \subset \mathbb{R}^{|V| \times |F_v|}$
and edge features $X_e \subset \mathbb{R}^{|E| \times |F_e|}$ to graph-, node-, or edge-level quantities of interest.
This mapping is mainly enabled through \textit{message passing} and \textit{aggregation}. 

Message passing refers to the process of passing information between the nodes and edges in a graph.
In each iteration, each node and edge receives a message
from their neighboring nodes $\mathcal{N}(v)$ and edges $\mathcal{E}(v)$,
which contains information about their local neighborhood.
The received messages are then aggregated to update the node and edge feature representations.

Aggregation is usually done by a differentiable operation, such as summation, averaging, or attention-weighted aggregation.
The process of message passing and aggregation is repeated multiple times,
where each iteration updates the node and edge feature representations
based on the features of their neighbors in the previous iteration.
Specifically, at step $k$ (or in the $k$-th layer), for each node $v$ we aggregate the features of its neighbors,
denoted by $\mathbf{x}^k_u \in \mathbb{R}^{1 \times d^k_n}$, as:
\begin{equation}
\mathbf{x}^{k+1}_{\mathcal{N} (v)} = f \big(\left\{ \mathbf{x}^k_u \right\}_{u \in \mathcal{N}(v)}\big), \quad \forall v \in V,
\end{equation}
where $f$ is an aggregation function and $d^k_n$ is the node embedding dimension at layer $k$.
Similarly, at step $k$, for each node $v$ we aggregate the features of its edges,
denoted by $\mathbf{x}_e^k \in \mathbb{R}^{1 \times d^k_e}$, as:
\begin{equation}
\mathbf{x}^{k+1}_{\mathcal{E}(v)} = g \big(\left\{ \mathbf{x}^k_e \right\}_{e \in \mathcal{E}(v)}\big), \quad \forall v \in V,
\end{equation}
where $g$ is an aggregation function and $d^k_e$ is the edge embedding dimension at layer $k$.
Then, the node features are updated using the previous features and the aggregated features as:
\begin{equation}
\mathbf{x}^{k+1}_v= \sigma(\mathbf{W}^k_v \mathbf{q}^k_v), \quad
\mathbf{q}^k_v = \left(\mathbf{x}^k_v,\mathbf{x}^{k+1}_{\mathcal{E}(v)}, \mathbf{x}^{k+1}_{\mathcal{N}(v)}\right)^\intercal,
\quad
\forall v \in V,
\end{equation}
where $\sigma$ is a nonlinear activation function and $\mathbf{W}^k_v$ is the parameter matrix of the $k$-th layer.

By repeatedly performing message passing and aggregation,
GNNs propagate information across the graph to capture complex dependencies among nodes and edges,
producing effective representations for tasks like prediction and classification at node, link, and graph levels.

\subsubsection{Applications of Multi-fidelity GNNs}

Like any other machine learning algorithm, GNNs are data-hungry, meaning that they require a large training set.
This becomes problematic when aiming to replace computationally expensive simulations of the systems with GNNs,
as generating a large training set by running high-fidelity simulations may not be computationally feasible.
This challenge can be alleviated using MF modeling.
However, the advantages of MF approaches in GNNs have remained relatively unexplored, and the literature on MF GNNs is scarce.

Rather than focusing on methodological advancements to make the MF training of GNNs more efficient,
the literature contains sparse applications of MF GNNs across different fields.
For example, \citet{black2022learning} introduced an MF GNN approach designed to improve physics-based modeling
in elastostatic problems by utilizing innovative low-fidelity projections and the abstraction of subdomains into subgraphs.
However, this method is predominantly restricted to two-dimensional settings with uniform grids,
which limits its applicability to more complex, higher-dimensional problems.
In another study, \citet{li2023multi} presented an MF GNN model for predicting flow fields in turbomachinery,
with fidelity levels determined by the simplification of physical equations and convergence criteria.
This approach, however, is not applicable to cases where fidelity is defined by mesh resolution,
which is a common practice in many engineering and scientific simulations.

In a recent study, \citet{taghizadeh2024multifidelity} introduced a hierarchical MF GNNs for mesh-based simulation of PDEs,
where the high-fidelity GNN estimates the discrepancy between PDE estimations using coarse and fine meshes.
The efficacy of the approach is demonstrated across a variety of experiments,
ranging from 2D stress analysis in structural components with changing designs to complex 3D CFD simulations.

MF GNNs have also been applied in the context of power systems.
For example, \citet{Taghizadeh2024111014} trained an MF GNN as a surrogate for power flow simulation,
using used AC and DC power flow simulations as HFM and LFM, respectively.
\citet{khayambashi2024hybrid} used the power flow surrogate developed by \citet{Taghizadeh2024111014}
to facilitate the solution of optimal power flow under demand and generation uncertainty.
Both works have demonstrated the robustness of their results to changes in system topology
(e.g., failure of lines in power systems),
which makes GNNs distinct from analytical or conventional machine learning surrogates. 

\subsection{Future Directions of Multi-fidelity Surrogates}

As the literature shifts from analytical surrogates toward machine learning-based surrogates,
studies appear to become more application-focused rather than method-oriented.
For example, various methods proposed for enhancing (multi-fidelity) PCEs are often tested on well-known benchmarks.
In contrast, works on MF GNNs tend to focus on specific applications.
This is because machine learning-based surrogates, particularly GNNs,
are highly adaptable to the nuances of different problem domains,
making them well-suited for application-specific challenges.

Still, the literature could significantly benefit from integrating concepts that have proven effective
for analytical surrogates, such as dimension reduction and optimal sampling,
into the training of machine learning-based surrogates.
One challenge in this integration is the multi-output nature of several machine learning-based surrogates,
which requires non-trivial extensions of such methods to be integrated with machine learning algorithms like GNNs. 

For instance, an MF GNN could potentially benefit from
an extension of the optimal sampling proposed by \citet{alemazkoor2022multi}.
This could be achieved by searching a large pool of sample candidates, $P$,
and their low-fidelity approximations to select the most significant samples for training the high-fidelity GNN.
One possible strategy is defining the most significant sample $w_{k}$ at the $k$-th iteration as:
\begin{equation} \label{eq:opt-sam} 
    w_{k}= \underset{w_{k} \in P}{\arg\max} \left [\text{dist}\left(f_L(w_k), F_L(\gamma^{k-1})\right) + \alpha \left \|f_L(w_k) -\widetilde{f}_L(w_k) \right \| \right ],
\end{equation}
where $f_L(w_k)$ and $\widetilde{f}_L(w_k)$ are the outputs of the low-fidelity simulation at $w_{k}$
and its approximation by the low-fidelity GNN, respectively.
Here, $\alpha$ is a weight parameter, $\gamma^{k-1}$ represents the set of selected samples in previous iterations,
and $F_L(\gamma^{k-1})$ depicts the vector space spanned by those samples.
In simpler terms, the first term in \cref{eq:opt-sam} selects, at each iteration, the point that maximizes the distance
from the corresponding multi-dimensional output to the vector space spanned by the outputs of the previously selected points.
The second term selects the point that corresponds to the largest approximation loss.

The benefit of such an extension of optimal sampling may vary across different applications
and depend on the quality of the LFM.
In other words, a single application of this method using specific models and data may not provide meaningful insights.
Therefore, it is crucial for the uncertainty quantification community
to acknowledge the growing importance of machine learning-based surrogates
and to actively facilitate their integration into broader UQ frameworks.
This includes applying these novel machine learning-based surrogates to established UQ benchmarks
and potentially creating new benchmarks to better challenge and assess the capabilities of machine learning-based surrogates.
By doing so, the community can ensure that these advanced methods are not just beneficial for specific applications or datasets
but have a wide range of applicability and are practically robust.
This would pave the way for more reliable and efficient machine learning-based UQ practices
across various scientific and engineering disciplines.

\section{Multi-fidelity Optimization and Design}
\label{sec:mfo}

Numerical optimization is crucial for a wide range of tasks in computational science and engineering,
including design optimization \citep{Do2023mfbo}, system identification, inverse problems, experimental design,
and machine learning model training.

Consider the following optimization problem:
\begin{equation}\label{eq:opt}
	\begin{aligned}
		\underset{\bf x}{\min} \ \ & f(\bf x)\\
		\textrm{subject to} \ \ 
		& \bf x \in \mathcal{X}, 
	\end{aligned}
\end{equation} 
where input vector ${\bf x} \in \mathbb{R}^d$ consists of $d$ variables in a bounded domain $\mathcal{X} \subset \mathbb{R}^d$,
and objective function $f({\bf x}):\mathbb{R}^d \mapsto \mathbb{R}$ is to be minimized.
The objective function is a mathematical abstraction that may represent physical experiments,
solutions of differential equations, or model predictions on large data sets.
As a result, evaluating the objective function can be too costly to allow for an adequate search for optimal solutions.

To improve data efficiency of the optimization process,
the objective function $f({\bf x})$ can be approximated with a surrogate model.
This surrogate may approximate the objective function either locally (e.g., Taylor series expansion)
or globally (e.g., response surfaces, radial basis functions, or ML models).
For example, given HF data in the form of input variables paired with objective values,
where the input space is adequately sampled,
one can train the parameters of a neural network to approximate the objective function globally.

Naively applying surrogate-based optimization may require a large number of HF data,
which defeats the goal of improving data efficiency.
\textit{Multi-fidelity optimization} (MFO) leverages both LF models and HF data in constructing surrogates.
Rather than using a generic function space for the approximation,
LFMs are used to encode a function space where good approximations are likely to exist,
which reduces the amount of HF data needed compared to purely data-driven surrogates.
Specifically, suppose there is an HFM $f_{\text{H}}(\mathbf{x})$ and an LFM $f_{\text{L}}(\mathbf{x})$,
MF surrogates can be constructed by:
additive adjustments, $f_{\text{H}}(\mathbf{x}) = a + b f_{\text{L}}(\mathbf{x}) + \delta(\mathbf{x})$;
multiplicative adjustments, $f_{\text{H}}(\mathbf{x}) = \rho(\mathbf{x}) f_{\text{L}}(\mathbf{x})$;
composition in the input space, $f_{\text{H}}(\mathbf{x}) = f_{\text{L}}(g(\mathbf{x}))$;
composition in output space, $f_{\text{H}}(\mathbf{x}) = h(f_{\text{L}}(\mathbf{x}))$;
or augmentation of the input with continuous or categorical fidelity labels, $f(\mathbf{x}, t)$.

While MFO exploits useful information in LFMs,
another approach to data-efficient surrogate-based optimization is \textit{Bayesian optimization} (BO).
Bayesian optimization is built on two key concepts: probabilistic function ensembles and Bayesian decision theory.
Instead of constructing deterministic surrogates for the deterministic objective function,
BO constructs an ensemble of functions that approximate the objective function,
in the form of a probability distribution on a function space.
The initial probabilistic model---known as the prior in Bayesian statistics---is updated
with the data set at each stage using the Bayes' rule.
In optimization, Bayesian decision theory provides a general framework for defining optimization policies
that account for such uncertainties in our belief about the objective function.
The next observation location is selected as the global maximal point of an acquisition function,
which is the expected utility function under the updated probabilistic model---also known as the posterior.

BO is a popular approach to surrogate-based optimization due to its many desirable properties:
\begin{enumerate*}[label={(\arabic*)}]
\item data efficiency in optimizing expensive-to-evaluate objective functions;
\item no requirement on derivative information;
\item robustness to noise;
\item flexibility in balancing exploitation and exploration; and
\item strong theoretical guarantees on convergence.
\end{enumerate*}
As an example, BO can be applied to system identification
to estimate the constitutive parameters for the web and the flange of a cantilever beam,
given experimental data under cyclic loading.
In \citet{Do2024egts}, ten parameters from a nonlinear isotropic–kinematic hardening model of steel
are estimated to minimize the root mean squared error (RMSE) in moment reaction.
A BO algorithm is used and it identifies models that accurately match the experimental data within 100 iterations.

Given the complementary nature of MFO and BO,
recent developments of MFO have largely focused on the intersection of these two approaches,
known as multi-fidelity Bayesian optimization (MFBO).
A recent review on this subject is provided in \citet{Do2023mfbo}.
For the remaining parts of this section, we therefore focus our discussion on MFBO.

\subsection{Key Elements of Multi-fidelity Bayesian optimization}
\label{sub:MFBO}

A Bayesian optimization algorithm is largely determined by three key components:
(1) a prior $\pi$ for the objective function,
(2) an optimization policy $\mathbb{P}$ that determines the next observation point given the posterior, and
(3) a global optimization program $\mathfrak{m}$ that maximizes the acquisition function numerically.

Most of the time, the prior model in BO is a Gaussian process (GP) due to its tractability and flexibility.
GP is a popular tool in probabilistic machine learning
that generalizes Gaussian distributions from Euclidean spaces to function spaces.
A GP model is specified by a mean function $m(\mathbf{x})$ and a covariance function $\kappa(\mathbf{x}, \mathbf{x}')$,
and can be written as $f \sim \mathcal{GP}(m, \kappa)$.
Although GP models are highly flexible, in BO typically the mean function is set to zero
and the covariance function is parameterized as the isotropic squared exponential (SE) kernel,
also known as the automatic relevance determination kernel:
$\kappa_{\text{SE}}(\mathbf{x}, \mathbf{x}'; \boldsymbol{\gamma}) = \exp\{-\frac{1}{2} \sum_{i=1}^d [(x_i - x'_i)/\gamma_i]^2\}$,
where $\gamma_i \in (0, \infty)$ is the length-scale in the $i$-th direction.
Such a prior guarantees the smoothness of the sample functions,
which is helpful since the objective function is likely to be smooth,
but it captures little information otherwise.
A key element in MFBO is a multi-fidelity prior,
where scientific knowledge is exploited in the form of low-fidelity models.
As with all MFO methods, MF priors in MFBO can be built from LFMs by adjustments, composition, or input augmentation.
These MF priors are much more structured than the prior $\mathcal{GP}(0, \kappa_{\text{SE}})$ used in generic BO.

The optimization policy can be specified either as a utility function $u(\mathbf{x}; f|\mathcal{D})$
or as an acquisition function $\alpha(\mathbf{x}; \mathcal{D})$.
Here, $\mathcal{D}$ denotes the current data set and $f|\mathcal{D}$ denotes the posterior model of the objective function.
The acquisition function is a real-valued function and the utility function is a random process;
they are related by $\alpha(\mathbf{x}; \mathcal{D}) = \mathbb{E}[u(\mathbf{x}; f|\mathcal{D})]$.
Common acquisition functions for BO include expected improvement (EI),
upper confidence bound (UCB), and Thompson sampling (TS) \citep{Adebiyi2024bdu}.
Another key element in MFBO is a multi-fidelity optimization policy,
which allows for adaptive evaluations of HFM and LFMs.
MF optimization policies often build on generic acquisition functions
and include considerations of fidelity and cost.
For example, MF optimization policies may select the observation location $\mathbf{x}$ and the fidelity level $t$ jointly,
by using an acquisition function of the form $\alpha(\mathbf{x}, t)$,
or select the fidelity level after the observation location is determined,
by using an acquisition function of the form $\beta(t; \mathbf{x})$.
Considerations for cost may be included as $\alpha(\mathbf{x}, t) = \alpha(\mathbf{x}) / c(t)$,
where $c(t)$ is the cost of evaluating the model $f_t$.
Considerations for fidelity may be included as
$\alpha(\mathbf{x}, t) = \alpha(\mathbf{x}) \, \text{Cor}[f_t(\mathbf{x}), f_{\text{H}}(\mathbf{x}) | \mathcal{D}]$,
where $\text{Cor}$ denotes the correlation coefficient between two random variables.

MFBO algorithms have seen a significant number of applications, especially in engineering design.
Compared with traditional BO algorithms, MFBO can often reduce the number of HFM evaluations by two or more times,
which is significant especially when the high-fidelity data comes from physical experiments.

\subsection{MFBO from the Lens of BO: Low-fidelity Information in the Prior}
\label{sub:LFM-prior}

While MFBO methods have been highly successful,
a unifying framework to explain their data efficiency improvement over generic BO methods is still lacking."
In other words, one may wonder: where does the low-fidelity information go?
With the key elements of MFBO highlighted in \cref{sub:MFBO},
a tempting answer is that MFBO provides (1) stronger priors and (2) smarter policies.
Here we provide a detailed explanation of the first point:
MFBO embeds low-fidelity information in the prior of the objective function.

\subsubsection{Adjustment MF Priors}

Since there are multiple types of MF priors, we will start with adjustment priors, which are the most common approach.
All additive adjustment GP priors can be written as a multi-output GP via the linear model of coregionalization (LMC).
Let $\mathbf{f}(\mathbf{x}) = (f_i(\mathbf{x}))_{i=1}^T$ be the outputs of $T$ models with different fidelities.
The LMC model can be written as $\mathbf{f}(\mathbf{x}) = \mathbf{m}(\mathbf{x}) + \mathbf{R} \boldsymbol{\delta}(\mathbf{x})$,
where $\mathbf{m}(\mathbf{x})$ are deterministic mean functions,
$\boldsymbol{\delta} \sim \mathcal{GP}(0, \diag(\kappa_i)_{i=1}^T)$ are independent zero-mean GPs,
and $\mathbf{R}$ is a matrix.
This results in a joint prior on the outputs:
\begin{equation}
    \mathbf{f} \sim \mathcal{GP}(\mathbf{m}, \mathbf{R} \diag(\kappa_i)_{i=1}^T \mathbf{R}^\intercal),
\end{equation}
where $\mathbf{R}$ affects the correlation among the models.

A special case of the LMC is the auto-regressive model by \citet{Kennedy2000},
where the fidelities of the models are assumed to form a hierarchy,
and each model is directly dependent only on the immediately lower-fidelity model:
\begin{equation}
    f_1(\mathbf{x}) = \delta_1(\mathbf{x}), \quad
    f_t(\mathbf{x}) = \rho_t f_{t-1}(\mathbf{x}) + \delta_t(\mathbf{x}), \quad
    t = 2,\cdots,T.
\end{equation}
This dramatically reduces the number of additional hyperparameters in the prior,
apart from those in the independent GPs $\boldsymbol{\delta}(\mathbf{x})$:
the correlation matrix $\mathbf{R}$ is $\frac{T (T-1)}{2}$-dimensional,
while the correlation coefficients $(\rho_t)_{t=2}^T$ are $(T-1)$-dimensional.

While there are a plethora of variants to the auto-regressive model,
we focus our discussion on the recursive model of \citet{Gratiet2014},
due to its expressivity and computational efficiency.
The recursive model is a hybrid additive--multiplicative adjustment model,
where each model is directly dependent on the posterior of the immediately lower-fidelity model:
\begin{equation}\label{eq:recursive-model}
    f_1(\mathbf{x}) = \delta_1(\mathbf{x}), \quad
    f_t(\mathbf{x}) = \rho_t(\mathbf{x}) \widehat{f}_{t-1}(\mathbf{x}) + \delta_t(\mathbf{x}), \quad
    t = 2,\cdots,T.
\end{equation}
Here, the multiplicative adjustments $(\rho_t(\mathbf{x}))_{t=2}^T = \mathbf{B} \, \boldsymbol{\zeta}_t(\mathbf{x})$
are linear combinations of deterministic basis functions,
the additive adjustments $(\delta_t)_{t=1}^T \sim \mathcal{GP}(\mathbf{m}_\delta, \diag(\boldsymbol{\kappa}_\delta))$
are independent GPs,
and $\widehat{f}_{t-1}(\mathbf{x})$ is the posterior on the LFM $f_{t-1}(\mathbf{x})$
constructed from data with fidelity levels up to $t-1$.
The use of posteriors in the recursive model decouples training across the fidelity levels,
which reduces the computational cost from $\mathcal{O}\big((\sum_{t=1}^T N_t)^3\big)$
to $\mathcal{O}\left(T \times \max\{N_t^3\}_{t=1}^T\right)$,
where $N_t$ is the number of data points at fidelity level $t$.

Denote the posterior as $\widehat{f}_{t-1}(\mathbf{x}) \sim \mathcal{GP}(\widehat{m}_{f,t-1}, \widehat{\kappa}_{f,t-1})$,
and define the kernel $\kappa_{\rho,t}(\mathbf{x}, \mathbf{x}') = \rho_t(\mathbf{x}) \rho_t(\mathbf{x}')$,
the prior for the model $f_t(\mathbf{x})$ can then be written as:
\begin{equation}\label{eq:LF-posterior-prior}
    f_t \sim \mathcal{GP}(\rho_t \widehat{m}_{f,t-1} + m_{\delta,t}, \;
    \kappa_{\rho,t} \widehat{\kappa}_{f,t-1} + \kappa_{\delta,t}).
\end{equation}
It is clear from \cref{eq:LF-posterior-prior} that the recursive model embeds low-fidelity information
into the prior of the next fidelity level.
In particular, the low-fidelity information is presented as a GP posterior given low-fidelity data.

It is also possible to embed LF models, rather than LF data, in an MF prior.
For simplicity, consider the situation with one HFM $f_{\text{H}}(\mathbf{x})$ and one LFM $f_{\text{L}}(\mathbf{x})$.
Suppose the LFM is known analytically. In practice, this assumption can be weakened
as long as the LFM allows fast evaluations of its values and gradients.
We model the HFM with a hybrid adjustment prior:
\begin{equation}
    f_{\text{H}}(\mathbf{x}) = \rho(\mathbf{x}) f_{\text{L}}(\mathbf{x}) + \delta(\mathbf{x}),
\end{equation}
where $\rho \sim \mathcal{GP}(m_\rho, \kappa_\rho)$ and
$\delta \sim \mathcal{GP}(m_\delta, \kappa_\delta)$ are independent.
Since this prior is a linear transform of the multi-output GP $(\rho(\mathbf{x}), \delta(\mathbf{x}))$,
the prior is also a GP, which can be written as:
\begin{equation}\label{eq:LFM-prior}
    f_{\text{H}} \sim \mathcal{GP}(m_\rho f_{\text{L}} + m_\delta, \; \kappa_\rho \kappa_{f,\text{L}} + \kappa_\delta),
\end{equation}
where $\kappa_{f,\text{L}}(\mathbf{x}, \mathbf{x}') = f_{\text{L}}(\mathbf{x}) f_{\text{L}}(\mathbf{x}')$.
It is therefore clear from \cref{eq:LFM-prior} that the LFM is directly embedded into the prior of this MF surrogate.
In cases where the LFMs are black-box or somewhat costly,
it can be more efficient to approximate them with GPs, as in the recursive model.

\subsubsection{Composition and Input-Augmentation MF Priors}

The second type of MF prior is based on composition.
The composition of Gaussian process models is also known as deep Gaussian processes (DGPs) which,
paradoxically, are not GPs in general.
The incorporation of LF information in composition MF priors depends on their specific construction.

The nonlinear auto-regressive model by \citet{Perdikaris2017} 
is similar to the recursive model in \cref{eq:recursive-model},
but it further imposes GP priors on the multiplicative adjustments and the LFM posteriors.
It can be written as:
\begin{equation}\label{eq:nonlinear-auto-regressive}
    f_1(\mathbf{x}) = \delta_1(\mathbf{x}), \quad
    f_t(\mathbf{x}) = \rho_t(\mathbf{x}) \, g_t(\widehat{f}_{t-1}(\mathbf{x})) + \delta_t(\mathbf{x}), \quad
    t = 2,\cdots,T,
\end{equation}
where $(\rho_t, g_t, \delta_t)_t$ are independent GPs.
Due to the composition and multiplication, the resulting prior on $f_t$ is no longer a GP,
but the low-fidelity information is still embedded as a posterior given low-fidelity data.

Multi-fidelity DGP models, such as the one in \citet{Cutajar2019}, can be written as:
\begin{equation}\label{eq:MF-DGP}
    f_1(\mathbf{x}) = \delta_1(\mathbf{x}), \quad
    f_t(\mathbf{x}) = g_t(f_{t-1}(\mathbf{x})), \quad
    t = 2,\cdots,T,
\end{equation}
where $(g_t, \delta_1)_t$ are independent GPs.
In this case, the prior on $f_t$ is again not a GP,
and the LF information is embedded in the prior through function values: $f_t \sim g_t \circ f_{t-1}$.

The third and final type of MF prior is based on input augmentation.
Let $\mathbf{t}$ be a vector of continuous or categorical labels of model fidelity.
Input-augmentation MF priors model a function on the extended input space of $\mathbf{x}$ and $\mathbf{t}$:
\begin{equation}
    g(\mathbf{x}, \mathbf{t}) \sim \pi,
\end{equation}
where the prior $\pi$ is often a GP.
Low-fidelity data, or models, are provided to the extended function as
$f_{\mathbf{t}}(\mathbf{x}) = g(\mathbf{x}, \mathbf{t})$.
The HFM is then predicted as a slice of the extended function:
$f_{\text{H}}(\mathbf{x}) = g(\mathbf{x}, \mathbf{t}_{\text{H}})$,
where $\mathbf{t}_{\text{H}}$ is the fidelity label of the HFM.
In input-augmentation methods, models are positioned ``side-by-side'',
and information is transferred between them via correlation.
Low-fidelity information is thus embedded in the prior of $f_{\text{H}}$
after conditioning $\pi$ on the LF data.

\subsection{A Practical Application: Quadrature Scheme Sparsification}
\label{sub:LF-quadrature}

Despite the many approaches to constructing MF priors, all assume that the models are correlated,
and higher correlation implies better data efficiency in the optimization process.
In fact, the fidelity of a model is often measured by its correlation with the ground truth,
rather than by some distance function.
This poses a dilemma for the applicability of MFBO:
in which problems do we find models that are cheap to evaluate yet highly correlated with the ground truth?
Here we point out a practical application scenario---sparse evaluation of quadrature schemes.

In design optimization, practitioners are often interested in the performance of a product under multiple operating conditions,
and the overall objective is specified as a weighted sum of the condition-specific performances:
\begin{equation}\label{eq:weighted-sum}
    f(\mathbf{x}) = \sum_{i=1}^n w_i \, g(\mathbf{x}, \boldsymbol{\xi}_i),
\end{equation}
where $g(\mathbf{x}, \boldsymbol{\xi}_i)$ is the performance of a design $\mathbf{x}$
under operating conditions $\boldsymbol{\xi}_i$,
and weights $w_i > 0$ are associated with the operating conditions and are independent of the design.

A related scenario is when the objective function is an integral over a given domain:
\begin{equation}\label{eq:integral}
    f(\mathbf{x}) = \int_\Omega g(\mathbf{x}, \boldsymbol{\xi}) \, \mu(\diff \boldsymbol{\xi}),
\end{equation}
where $\boldsymbol{\xi} \in \Omega \subset \mathbb{R}^s$ and $\mu$ is a measure on the domain $\Omega$.
Such an integral may arise, for example, as a continuous version of \cref{eq:weighted-sum},
where $\Omega$ is the space of possible operating conditions and $\mu$ is the weight assigned to each operating condition.

Integrals also arise as expectations.
In optimization under uncertainty, the objective function is an integral with respect to the uncertain elements:
$f(\mathbf{x}) = \mathbb{E}[y|\mathbf{x}]$, where $y$ is the observation.
The uncertain elements can be of various natures, including:
\begin{enumerate*}[label={(\arabic*)}]
\item noisy observations, with $g(\mathbf{x}, \xi) = h(\mathbf{x}) + \xi$;
\item perturbed designs, with $g(\mathbf{x}, \boldsymbol{\xi}) = h(\mathbf{x} + \boldsymbol{\xi})$; or
\item uncertain parameters, with a general $g(\mathbf{x}, \boldsymbol{\xi})$.
\end{enumerate*}
A related problem is robust optimization, where integrals arise if the objective function
is a moment involving variance \citep{Kanno2020},
i.e., $f(\mathbf{x}) = \mathbb{E}[s(y)|\mathbf{x}]$ where $s$ is a nonlinear function,
or if it is the conditional value-at-risk \citep{Cakmak2020},
$f(\mathbf{x}) = \mathbb{E}[y|\mathbf{x}, y > \text{VaR}(\mathbf{x}; \pi)]$,
where value-at-risk $\text{VaR}(\mathbf{x}; \pi)$ is the $\pi$-quantile of $y|\mathbf{x}$.

In all scenarios, the integral in \cref{eq:integral} can be estimated by a quadrature scheme in the form of \cref{eq:weighted-sum},
where $\boldsymbol{\xi}_i$ is called a node and $w_i$ is called the corresponding weight.
Common quadrature schemes in one dimension include
Clenshaw--Curtis, Gauss--Legendre, and Gauss--Hermite quadrature \citep{Trefethen2022};
for multiple dimensions, popular methods include
tensor product, sparse grid, adaptive quadrature, and quasi-Monte Carlo methods \citep{Dick2013}.

Quadrature schemes are particularly useful in evaluating integrals of expensive-to-evaluate functions,
because they are designed to minimize approximation error for any given number of nodes.
For example, in designing a photonic energy device using BO,
\citet{Do2024photonics} implemented a tensor product rule based on transformed Gauss quadrature
to evaluate the objective function, which is an integral over a rectangular domain.
Compared with an equal-spaced rectangular grid,
the quadrature scheme achieves an order of magnitude speedup,
while reducing relative error from $\mathcal{O}(10^{-3})$ to $\mathcal{O}(10^{-7})$.

With the additive structure of quadrature schemes, it is very easy to construct LFMs.
Essentially, every subset of the quadrature terms forms an LFM:
because of the positivity of the quadrature weights,
for any index set $I \subset \{1, \cdots, n\}$,
$f_I(\mathbf{x}) = \sum_{i \in I} w_i \, g(\mathbf{x}, \boldsymbol{\xi}_i)$
is (positively) correlated with the quadrature scheme in \cref{eq:weighted-sum}.
Notice that $f_I(\mathbf{x})$ can incur large approximation error in the integral,
so it is typically not a good quadrature scheme by itself,
but due to its correlation with the full quadrature scheme,
it works well as an LFM for the objective function in \cref{eq:integral}.
While it is possible that better correlations can be achieved by updating the weights $\{w_i\}_{i \in I}$,
the reweighting should be carefully chosen together with the subset $I$.
An example application of sparsified quadrature schemes in aerospace design is provided in \citet{Do2023mfbo},
where an LFM is defined by evaluating only one of the multiple operating conditions considered in the objective function.

To further exploit the advantage of MFBO, one needs more than one LFM.
A natural way to construct multiple sparsified quadrature schemes while retaining computational efficiency
is to use nested quadrature rules.
Nested quadrature rules use a hierarchy of quadrature points,
where points from coarser levels are reused at finer levels.
The nodes and weights at each level of the hierarchy thus constitute an LFM,
and the LFMs naturally form a chain of increasing fidelity.
The sequential order of model fidelity is well suited for many MFBO methods,
such as the recursive model \citep{Gratiet2014} and continuous approximations \citep{Kandasamy2017}.
Depending on the convergence rate of the quadrature scheme used,
we can expect to reduce the number of integrand evaluations by one or a few orders of magnitude.

\subsection{Challenges and Possible Directions in MFBO}
\label{sub:MFBO-challenges}

MFBO has been successful in a wide range of applications, and its many modeling elements make it highly customizable.
Here we outline a few challenges that MFBO still faces.

\textit{Tighter integration of scientific knowledge in BO, beyond black-box models.}
BO treats the objective function as a black box, ignoring the rich structures in scientific applications.
In comparison, MFBO uses LFMs that are cheap to evaluate to accelerate BO.
While scientific knowledge is exploited in the form of LFMs,
the current practice of MFBO essentially breaks one black box into many small black boxes.
To further utilize our scientific knowledge, we need to delve into the structure of design optimization problems,
such as design parameterization, physical processes, performance metrics, and preferences over outcomes \citep{Do2023mfbo}.
A tighter integration means designing MFBO algorithms that preserve significant aspects of this structure.
There have been some attempts in this direction.
For example, our discussion in \cref{sub:LF-quadrature} can be seen as
exploiting the known structure of integrals and weighted sums in the objective function.
This falls into the category of utilizing our preferences on performance.
An example of leveraging design parameterization to accelerate the optimization process can be found in \citet{Do2024photonics}.
Perhaps the more interesting question is how we can leverage our knowledge of the physical processes to design stronger priors.
Following our discussion in \cref{sub:LFM-prior},
current MFBO methods can be seen as building stronger priors by embedding LF data,
which impacts the prior locally.
Embedding knowledge of the physical processes opens up the possibility of impacting the prior globally,
which may significantly improve data efficiency.
One way to build a physics-informed GP prior is to 
embed the governing equations into the covariance function (see, e.g., \citep{Alvarez2013}),
but this is generally not applicable to nonlinear systems.

\textit{Principled derivation of MF optimization policy.}
Currently, most optimization policies that consider model fidelity are based on heuristics.
For a joint selection of the observation location $\mathbf{x}$ and the fidelity level $t$,
MF acquisition functions are usually of the form $\alpha(\mathbf{x}, t) = \alpha(\mathbf{x}) \beta(\mathbf{x}, t)$,
where $\alpha(\mathbf{x})$ is an acquisition function for generic BO,
and $\beta(\mathbf{x}, t)$ is a multiplicative adjustment that represents
cost savings and/or information gain (e.g., correlation or uncertainty reduction in predictions)
associated with each model.
When the fidelity level $t$ is selected after the observation location $\mathbf{x}$,
a separate function $\gamma(t; \mathbf{x})$ is used,
constructed similarly to $\beta(\mathbf{x}, t)$.
Such constructions of MF optimization policies are crude in the sense that they are not justified by Bayesian decision theory,
which specifies a utility function $u(\mathbf{x}; f|\mathcal{D})$ that measures the quality of an action,
and derives the acquisition function as the expected utility:
$\alpha(\mathbf{x}; \mathcal{D}) = \mathbb{E}[u(\mathbf{x}; f|\mathcal{D})]$.
A more principled derivation may bring some of the current MF optimization policies closer to Bayesian decision theory,
and could lead to the discovery of more effective optimization policies for MFBO.

\textit{Global optimization of acquisition function.}
While most of the literature in BO and MFBO addresses the first two key components outlined in \cref{sub:MFBO},
the third key component of BO---the global optimization program $\mathfrak{m}$---is often overlooked.
When the acquisition function $\alpha(\mathbf{x})$ is known analytically and has a limited number of local optima,
it is relatively easy to optimize and does not pose much difficulty.
One may use, for example, a gradient-based local optimizer with multiple starting points.
This often applies to acquisition functions in generic BO that are averages over stochastic elements,
such as expected improvement (EI) or upper confidence bound (UCB).
When $\alpha(\mathbf{x})$ involves a sample function from the posterior $f|\mathcal{D}$,
such as Thompson sampling (TS) or information-theoretic acquisition functions, the optimization becomes much more challenging
because posterior samples have a large number of local optima that grows exponentially with the dimension $d$ of the design space.
To mitigate this challenge, \citet{Adebiyi2024roots} proposed an efficient strategy that is applicable to common GP priors.
For MFBO, all these acquisition functions become more difficult to optimize,
because MFBO incorporates low-fidelity models and data into the prior,
resulting in more local optima in the acquisition functions.
Therefore, efficient global optimization strategies must be carefully formulated for acquisition functions in MFBO,
even for those well studied in generic BO such as EI and UCB.

\textit{Convergence results for MFBO.}
Many BO algorithms are supported by theoretical guarantees of convergence to the global optimum of the objective function,
but such guarantees are lacking for MFBO.
For BO, these results come in various forms:
upper bounds for specific algorithms and lower bounds for any algorithms;
on the (Bayesian) expected regret and (frequentist) worst-case regret;
with and without observation noise.
Specific BO algorithms with convergence guarantees use, for example,
a suitable GP prior (usually zero mean and a \matern covariance)
and the acquisition functions UCB \citep{Srinivas2010}, TS \citep{Russo2014}, or EI \citep{Bull2011}.
To bring more rigor to MFBO algorithms, it is desirable to develop a theory for their convergence.
Unique to MFBO, convergence results should perhaps include a measure of model fidelity, such as correlation of predictions.
The corresponding MF optimization policy may also carefully balance adaptivity
and independence from the observed values, for example, via an epsilon-greedy policy
that mixes an adaptive policy with a non-adaptive exploratory policy \citep{Do2024egts}.
Considering the technical difficulty of convergence proofs,
a more fruitful direction for MFBO may be to improve their convergence in practice.
A general rule of thumb is to use high-quality LFMs, and fall back to a generic prior
or a non-adaptive policy when data does not support the MF prior.

\section{Conclusion}
\label{sec:conclusion}

This paper has provided an overview of multi-fidelity machine learning methods,
with a focus on their applications in uncertainty quantification and optimization.
While these methods offer significant potential, challenges and gaps in the literature remain.
Our discussion highlights the need for continued research in several key areas,
including the development of more robust and scalable MF frameworks,
the exploration of novel machine learning architectures,
and the application of these methods to increasingly complex systems.
By identifying these opportunities, we hope to encourage further investigation and innovation in the field,
ultimately leading to more effective and practical solutions for real-world problems.

\section*{Acknowledgements}

RZ acknowledges the financial support from the University of Houston
under the High Priority Area Research Seed Grant No.~000189862.
NA Acknowledges funding from Center for Hardware and Embedded System Security and Trust (CHEST)
and Commonwealth Cyber Initiative (CCI).

\bibliographystyle{jmlmc-natbib-Last-F-no-id}

\bibliography{MFML}

\end{document}